\documentclass[10pt,twocolumn,letterpaper]{article}

\usepackage[pagenumbers]{cvpr}      

\usepackage{graphicx}
\usepackage{amsmath}
\usepackage{amssymb}
\usepackage{booktabs}
\usepackage{epsfig}
\usepackage{amsthm}
\usepackage[ruled,vlined]{algorithm2e}
\usepackage[mode=buildnew]{standalone}

\usepackage{bbm}
\usepackage{comment}
\usepackage{subcaption}

\newcommand{\bx}{\mathbf{x}}
\newcommand{\bz}{\mathbf{z}}
\newcommand{\by}{\mathbf{y}}
\newcommand{\bw}{\mathbf{w}}
\newcommand{\bdx}{\delta \mathbf{x}}
\newcommand{\bdy}{\delta \mathbf{y}}
\newcommand{\bdz}{\delta \mathbf{z}}

\DeclareMathOperator*{\argmax}{argmax}

\graphicspath{{./figures/}{./figures/livingroom/}}

\usepackage[pagebackref,breaklinks,colorlinks]{hyperref}

\usepackage[capitalize]{cleveref}
\crefname{section}{Sec.}{Secs.}
\Crefname{section}{Section}{Sections}
\Crefname{table}{Table}{Tables}
\crefname{table}{Tab.}{Tabs.}

\def\cvprPaperID{1336} 
\def\confName{CVPR}
\def\confYear{2022}

\begin{document}

\title{Rayleigh EigenDirections (REDs): \\ GAN latent space traversals for multidimensional features}

\author{Guha Balakrishnan\\
Rice University\\
{\tt\small guha@rice.edu}
\and
Raghudeep Gadde\\
Amazon\\
{\tt\small rggadde@amazon.com}

\and
Aleix Martinez\\
Amazon\\
{\tt\small maleix@amazon.com}

\and
Pietro Perona\\
Amazon Web Services\\
{\tt\small peronapp@amazon.com}
}

\maketitle

\begin{abstract}
We present a method for finding paths in a deep generative model's latent space that can maximally vary one set of image features while holding others constant. Crucially, unlike past traversal approaches, ours can manipulate multidimensional features of an image such as facial identity and pixels within a specified region. Our method is principled and conceptually simple: optimal traversal directions are chosen by maximizing differential changes to one feature set such that changes to another set are negligible. We show that this problem is nearly equivalent to one of Rayleigh quotient maximization, and provide a closed-form solution to it based on solving a generalized eigenvalue equation. We use repeated computations of the corresponding optimal directions, which we call Rayleigh EigenDirections (REDs), to generate appropriately curved paths in latent space. We empirically evaluate our method using StyleGAN2 on two image domains: faces and living rooms. We show that our method is capable of controlling various multidimensional features out of the scope of previous latent space traversal methods: face identity, spatial frequency bands, pixels within a region, and the appearance and position of an object. Our work suggests that a wealth of opportunities lies in the local analysis of the geometry and semantics of latent spaces.
\end{abstract}

\begin{figure}[ht!]
\vspace{0in}
    \centering
    \includegraphics[width=\linewidth]{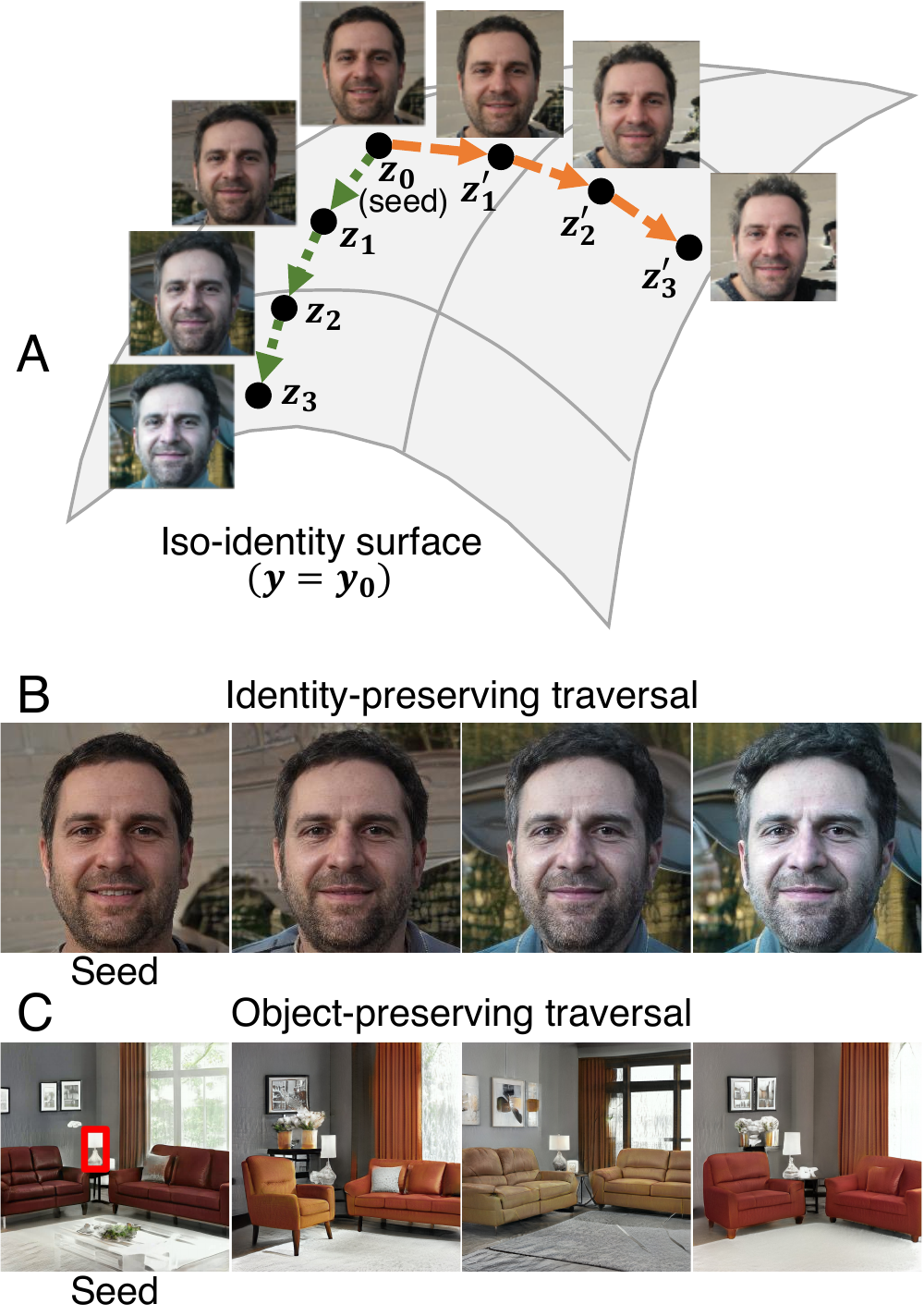}
    \caption{\textbf{Method and examples}. (A) Our method traverses the local latent space around a `seed point' $\bz_0$ along optimally chosen paths to synthesize images that share the same high-dimensional attribute value $\by_0$ (e.g., identity), and vary as much as possible across other image attributes (e.g., lighting, expression, age, hairstyle).  (B) Sample of an identity-preserving face traversal and (C) a living room traversal with a fixed object (lamp in red frame of seed image) in the latent space of StyleGAN2 generators.} 
    \label{fig:teaser}
\end{figure}


\section{Introduction}
Latent spaces of deep generative networks like generative adversarial networks (GANs)~\cite{goodfellow2014, karras2019style, karras2019analyzing,radford2015unsupervised} and variational autoencoders (VAEs)~\cite{Kingma2014} are known to organize semantic attributes into disentangled subspaces without supervision~\cite{ganspace, Jahanian2020On, radford2015unsupervised, upchurch2017deep, wang2021a}. This property is the basis of several \emph{traversal} algorithms proposed in the literature that can modify specific image attributes while holding others constant by moving along carefully-chosen latent space directions~\cite{balakrishnan2020towards,goetschalckx2019ganalyze,Plumerault2020Controlling,shen2019interpreting, yang2021semantic}. Traversal methods have many potential applications including dataset creation/augmentation, image editing, entertainment and graphic design.



Virtually all existing traversal methods assume \emph{scalar} attributes of interest that may be modeled well with global linear functions, e.g., a linear regressor or a support vector machine, in the latent space. This approach works well for attributes like gender, hair color and smile of faces~\cite{balakrishnan2020towards,shen2019interpreting} and image transformations like translation, color change and camera movements~\cite{Jahanian2020On,Plumerault2020Controlling}. But these approaches cannot be easily extended to work with attributes like `style of a couch' and `face identity' which are best described with high-dimensional vectors.\footnote{There is no individual person behind a GAN-generated portrait and therefore there is no physical `identity' ground truth. However, human observers or face recognition algorithms can respond to the question ``Is this the same person?'' and can produce consistent judgments. Therefore `identity' here denotes `perceptual identity'.} For example, to find a latent space traversal that preserves identity in our experiments, we need a representation that can compute the similarity between two $512$-dimensional embeddings returned by a face recognition model~\cite{deng2019arcface}. In addition, faces with the same identity or rooms with the same furniture layout (see Fig.~\ref{fig:teaser}C) tend to be tightly clustered in latent space, requiring methods tuned to local latent space geometry unlike the common global linear models used for scalar attributes.

We propose a method to tackle this broader class of traversal problems. Given a point in latent space, we aim to generate many traversals, or sequences of images, such that we vary one multidimensional feature ($\bx$) in as many ways as possible subject to other multidimensional features ($\by$) being held approximately constant. We formalize the task of finding local latent directions that fulfill these criteria as a constrained optimization problem. By using differential approximations of the feature functions, we recast the problem into an instance of Rayleigh quotient maximization, which has a well-known closed-form solution (Sec.~\ref{sec:differential}). The principal directions that solve this problem, which we call Rayleigh EigenDirections(REDs), span the local latent subspace containing good paths. Using REDs, we propose a fast linear and more accurate iterative nonlinear projection traversal algorithm (Sec.~\ref{sec:traversal}) to produce arbitrary-length paths. Our approach is agnostic to network architecture, scene content, and choice of attribute embedding functions.

We evaluate our method using StyleGAN2~\cite{karras2019style, karras2019analyzing} generators. We consider a number of challenging applications outside the scope of previous GAN traversal algorithms: face traversals that preserve identity (Fig.~\ref{fig:id-hair-landmark-samples}) while changing hairstyle and facial geometries, face traversals that preserve/change content from specific spatial frequency bands (Fig.~\ref{fig:freq-results}), and living room traversals that preserve the appearance and location of selected pieces of furniture (Fig.~\ref{fig:livingroomtraversal}). We provide a number of qualitative results demonstrating the perceptual quality of our generated image sequences, and quantitatively demonstrate the necessity for nonlinear traversal strategies in these applications. Finally, we also compare our method against well-known global linear model baselines~\cite{balakrishnan2020towards, shen2019interpreting} for scalar attributes and perform comparably, though with some failure cases that we discuss in Sec.~\ref{sec:limitations}.

Our main contributions are: (a) REDs, a {\em local} method for synthesizing a diverse set of images that share a chosen set of multidimensional attribute. The method is principled, simple, and versatile -- applicable to pretrained generators, any image type, and to both low-level and semantically meaningful features. (b) A nonlinear technique for long-distance traversals in latent space; (c) Qualitative and quantitative validation experiments on a number of challenging synthesis tasks in two different image domains.

\begin{figure*}[ht!]
    \centering
    \includegraphics[width=\linewidth]{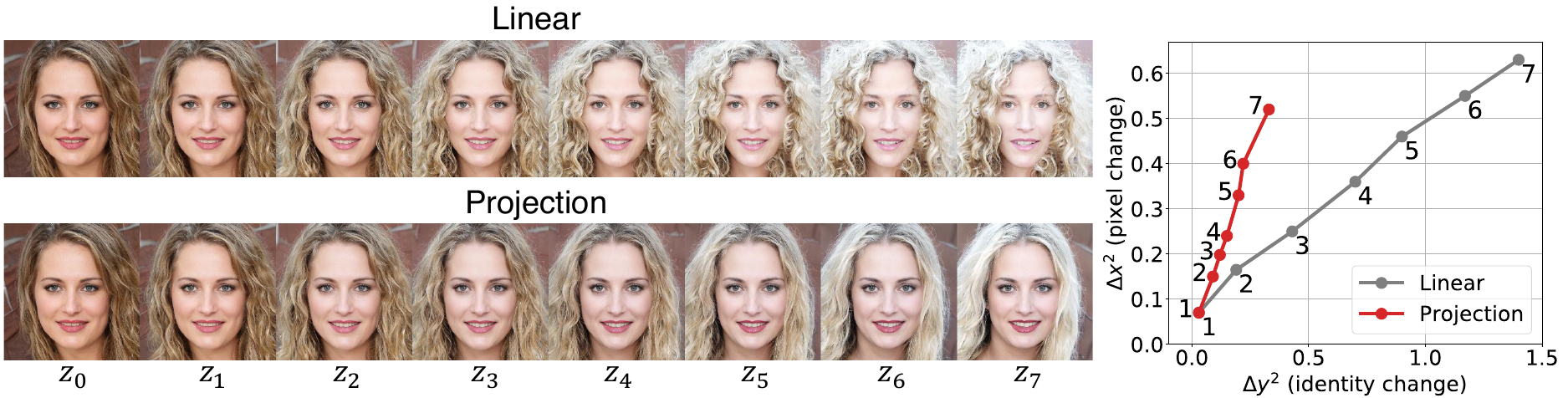}
    \caption{\textbf{Comparison of \emph{Linear} and \emph{Projection} traversal.} We show a \emph{Linear} and \emph{Projection} traversal originating from the same latent seed code (left-most face), and top RED vector at the seed. $f(\cdot)$ measures identity and $c(\cdot)$ measures raw face pixel values. We also plot squared pixel distance versus squared identity distance. \emph{Projection} and \emph{Linear} change pixels by roughly the same amount, but \emph{Projection} is better at preserving identity (lower distance values).}
    \label{fig:strip-example}
\end{figure*}

\section{Related Work}
\label{sec:related-work}
Several studies focus on finding interpretable directions in GAN latent spaces for editing and synthesizing images. Most propose finding global linear directions correlated with scalar attributes of interest~\cite{balakrishnan2020towards,ganspace,goetschalckx2019ganalyze,Plumerault2020Controlling,shen2019interpreting,voynov2020unsupervised, yang2021semantic}. One popular technique is to train a linear predictor, e.g., SVM, from latent codes and corresponding attribute labels, and use the norm of the learned hyperplane as the traversal direction~\cite{balakrishnan2020towards, shen2019interpreting}. We find that multidimensional features like face identity and hairstyle lie on complex manifolds in latent space rather than on simple linear ones, requiring locally-varying, nonlinear traversals. We propose an algorithm (see Sec.~\ref{sec:traversal}) that forms nonlinear paths by sequentially taking locally optimal linear steps. A few nonlinear traversal strategies do exist in the literature, typically based on training nonlinear neural networks to map latent codes to features~\cite{Jahanian2020On,tzelepis2021warpedganspace,yang2021discovering}. Our method is complementary to these -- ours requires no additional training, but also does not leverage global latent space structure as theirs presumably can. Finally, our focus on a local rather than global view of the latent space may also complement various theoretical studies on understanding GAN latent space structure~\cite{arvanitidis2017latent, balestriero2020max,chen2018metrics, kuhnel2018latent, shao2018riemannian,wang2021a}.


A more explicit way to control GAN outputs is to train the generator using attribute values as inputs. Many of these so-called ``conditional GANs'' have been proposed, particularly for altering face attributes~\cite{antipov2017face, bao2018towards, choi2018stargan, he2019attgan, kocaoglu2018causalgan, lample2017fader, liu2019stgan, mirza2014conditional, odena2017conditional, or2020lifespan, shoshan2021gan, tran2017disentangled, xiao2018elegant, yin2017towards}, controlling face identity~\cite{bao2018towards, shen2018faceid, shen2018facefeat, shoshan2021gan}, and conditioning on semantic maps~\cite{park2019semantic, wang2018high}. Our approach is complementary to all of these in that offers the benefit of not needing to design and train a GAN from scratch with apriori-known attribute controls. Working with a general-purpose black-box GAN has the advantage of keeping all control objectives open and not committing to a specific goal, e.g. preserving identity, from the beginning.

\section{Method}
\label{sec:methods}
Given a point \mbox{$\bz_0 \in \mathcal{R}^d$} in latent space defining an image, we want to generate a set of images that holds fixed the multidimensional features $\by_0 \in \mathcal{R}^n$ while maximally changing the features $\bx_0 \in \mathcal{R}^m$. 
For ease of explanation, we assume $\by_0$ and $\bx_0$ each define a single multidimensional feature like facial identity or hairstyle, though our method easily handles features from multiple semantic attributes as explained in Sec.~\ref{sec:multiattributes}. 

We denote the function that computes the {\em fixed} features \mbox{$f(\cdot): \bz \rightarrow \by \in \mathcal{R}^n$}, and the function that computes the {\em changing} features \mbox{$c(\cdot): \bz \rightarrow \bx \in \mathcal{R}^m$}. For example, in one of our experiments with faces, $f(\cdot)$ is the concatenation of two functions: the GAN generator on the input latent vector, and a face recognition embedding model on the synthesized face. $c(\cdot)$ may be the generator itself (i.e., $\bx$ are the raw pixels of the image) or the concatenation of the generator with learning models computing various image attributes.


Starting at \mbox{$\bz_0$}, our method traverses different paths in latent space to generate latent code sequences. For each such trajectory $t$ of length $L$, \mbox{$\bz_0, \bz^t_1 \cdots, \bz^t_L$}, we want \mbox{$\by^t_i \approx \by_0$} for all $i$ and \mbox{$\bx_0, \bx^t_1, \cdots, \bx^t_L$} to progressively change such that
\mbox{$\| \bx^t_i - \bx^t_{i+1} \| < \| \bx^t_i - \bx^t_{i+2} \|$}, where $\| \cdot \|$ is a norm. We return all points from all sequences. 

The key intuition behind our approach is that there exists a manifold on which $\by$ does not change around $\bz_0$  (see Fig.~\ref{fig:teaser}). This is true whenever $d>n$ (and thus the iso-$\by$ manifold has dimension $n-d$) and the generator function is continuous (which, by inspection, it is, apart from a zero-size set). When $d \leq n$, our approach naturally transitions to a ``soft'' constraint \mbox{$\by_i \approx \by_0$} as will become clear below. We find directions, which we call Rayleigh EigenDirections (REDs), that maximally change $\bx$ within this subspace. This procedure is described in Sec.~\ref{sec:differential}. We propose two traversal strategies using REDs in Sec.~\ref{sec:traversal}: a linear method which simply extrapolates the local REDs throughout the latent space, and a nonlinear method (\emph{Projection}) which updates traversal directions based on local latent space geometry.

\subsection{Rayleigh EigenDirections (REDs)}
\label{sec:differential}
Let $\bz$ be a generic point in the generator's latent space with fixed and changing features \mbox{$\by=f(\bz)$} and \mbox{$\bx=c(\bz)$}. Given a displacement $\bdz$, the displacements to $\by$ and $\bx$ are:
\begin{align}
\bdy &= f(\bz + \bdz) - f(\bz)\\
\bdx &= c(\bz + \bdz) - c(\bz).
\end{align}
We aim to find the displacement $\bdz^*$ that maximizes $\bdx$ with insignificant changes to $\bdy$:
\begin{align}
\bdz^* = & \argmax_{\bdz: \|\bdz\| = \epsilon}\| \bdx(\bz, \bdz) \|^2 \\
& \text{ s.t. } \| \bdy(\bz, \bdz) \|^2 \approx 0,
\end{align}
\noindent where we write $\bdx$ and $\bdy$ as functions of $\bz$ and $\bdz$, and $\epsilon$ is a small, fixed constant. For sufficiently small $\epsilon$, we can approximate $\bdy$ and $\bdx$ with local linear expansions: \mbox{$\bdy \approx J_f(\bz) \bdz$} and \mbox{$\bdx \approx J_c(\bz) \bdz$}, where \mbox{$J_f \in \mathcal{R}^{n \times d}$} and \mbox{$J_c \in \mathcal{R}^{m \times d}$} are Jacobian matrices. Letting \mbox{$A_f(\bz) = J^T_f(\bz)J_f(\bz)$} and \mbox{$A_c(\bz) = J^T_c(\bz)J_c(\bz)$}, we get:
\begin{align}
\bdz^* = & \argmax_{\bdz:\|\bdz\| = \epsilon} \bdz^TA_c(\bz)\bdz \label{eqn:obj} \\
& \text{ s.t. } \bdz^TA_f(\bz)\bdz \approx 0 \label{eqn:constraint}
\end{align}
\noindent This optimization is similar to one of finding the $\bdz$ that maximizes the Rayleigh quotient $\left(\bdz^TA_c(\bz)\bdz \right) / \left( \bdz^TA_f(\bz)\bdz \right)$, known to be the solution of the generalized eigenvalue problem $A_c \bdx = \lambda A_f \bdx$, or the principal eigenvector of $A_f^{-1}A_c$ (see Supplementary). The main point of difference is that in our applications $A_f$ is often singular ($n<d$) and therefore not invertible. Put another way, $f(\cdot)$ is constant in a subspace $\text{null}(A_f)$ around $\bz$ and any $\bdz$ in that subspace will exactly satisfy constraint~\eqref{eqn:constraint}. We instead first project $A_c$ onto $\text{null}(A_f)$, and then find the principal eigenvectors of the resulting matrix (Alg.~\ref{alg:reds})~\cite{ghojogh2019eigenvalue}. We return the top eigenvectors (REDs) in matrix $R \in \mathcal{R}^{d\times s}$, where $s$ is some integer, to define the local subspace of good traversal directions.

For some high-dimensional features, the rank of  $\text{null}(A_f)$ may be too small (or even 0 when $d < n$), yielding little to no diversity of $\bx$ in the generated trajectories. To address this, we introduce hyperparameter $\beta_f$ in Alg.~\ref{alg:reds} that lets users smoothly control the approximation of $A_f$'s rank based on explained variance. We also introduce $\beta_c$ to control the rank of the REDs matrix $R$.

The main computational cost of finding REDs is in calculating the Jacobian matrices $J_f$ and $J_c$. We compute them using two-sided finite difference approximations with step size $\epsilon$, which requires $2d+1$ forward evaluations of $f(\cdot)$ and $c(\cdot)$.

\begin{algorithm}[t!]
\textbf{Input}: $A_f$, $A_c$, $\beta_f$, $\beta_c$ \\
\textbf{Output}: $R$\\ 
\vspace{0.25cm}
$A_f, A_c \leftarrow A_f/\|A_f\|_2, A_c/\|A_c\|_2$\\
$\textbf{u}_f, V_f \leftarrow$ \text{eig}$(A_f)$\\
$\text{rank}_f \leftarrow \text{smallest } k \text{ s.t. }\sum_{i=0}^{k}{\textbf{u}_f^2(i)} \geq \beta_f\|\textbf{u}_f\|^2$ \\
$\text{null}(A_f) \leftarrow V_f[:, \text{rank}_f : d]$\\
$\tilde{\textbf{u}}_c, \tilde{V}_c \leftarrow$ \text{eig}$(\text{null}(A_f)^TA_c\text{ null}(A_f))$\\
$\text{rank}_c \leftarrow \text{smallest } k \text{ s.t. }
\sum_{i=0}^{k}{\tilde{\textbf{u}}_c(i)} \geq \beta_c\|\tilde{\textbf{u}}_c\|^2$ \\
$R \leftarrow \text{null}(A_f)\tilde{V}_c[:, 0 : \text{rank}_c]$
 \caption{Compute local REDs (solves optimization problem~\eqref{eqn:obj}-\eqref{eqn:constraint})}
 \label{alg:reds}
\end{algorithm}

\subsection{Fixing multiple attributes}
\label{sec:multiattributes}
In practice, we often want to fix multiple attributes simultaneously. One way to do so is to simply concatenate them together into $\by$. However, this approach offers limited individual control over each feature's variability.

Instead, given multiple features $\by^1, \cdots, \by^{n_f}$, we replace~\eqref{eqn:constraint} with multiple constraints: $\bdz^TA^i_f(\bz)\bdz \approx 0, i = 1 \cdots n_f$, and introduce a separate $\beta^i_f$ for computing the rank of each $A^i_f$. We compute REDs by projecting $A_c$ onto $\cap_{i=1}^{n_f}\text{null}(A^i_f)$ -- the intersection of the fixed attribute nullspaces -- and returning the top eigenvectors of the resulting matrix as before. 

\subsection{Traversal Algorithms}
\label{sec:traversal}

We propose two traversal algorithms using REDs. The first is a simple \emph{Linear} traversal (see Supplementary for algorithm). We randomly select a direction in the span of $R_0$ (the REDs of $\bz_0$), and generate a sequence of latent codes $\bz_1, \cdots , \bz_K$ by moving in that direction starting from $\bz_0$ with step size $s$. In the likely case that the constant-$\by$ manifold is curved, the linear traversal is expected to diverge quadratically from $\| \bdy \|=0$ as a function of $\|\bdz\|$.



Our second algorithm, \emph{Projection} (see Supplementary for algorithm), addresses this shortcoming by recomputing the space of local REDs along the traversal path. We again start by selecting a random direction in $R_0$. However, at each step $i$ (of length $s$), we project the previous direction, $\bdz_{i-1}$, onto $R_i$. This results in a path that more faithfully adheres to the local geometries of $f(\cdot)$ and $c(\cdot)$ in latent space.

A visual example of a \emph{Linear} and \emph{Projection} traversal for the same initial latent code is shown in Fig.~\ref{fig:strip-example}, where $f(\cdot)$ measures identity and $c(\cdot)$ measures raw face pixels. \emph{Projection} is better than \emph{Linear} at preserving identity for long trajectories (right plot), while achieving similar levels of image change (left plot).

\section{Experiments}
We evaluate our method on two image domains: faces and living rooms. We use StyleGAN2~\cite{karras2019analyzing} with \emph{config-f} configuration for both domains. For faces, we use the public model from NVIDIA trained on the Flickr Faces HQ (FFHQ) dataset~\footnote{https://github.com/NVlabs/stylegan2}. For living rooms, we train the GAN from scratch on an in-house dataset of 100K $1024 \times 1024$ living room scenes from the web. We use StyleGAN2's ``style'' space, $\bw \in \mathcal{R}^{512}$, as our latent space for both applications.

\begin{figure*}[t!]
    \centering
    \includegraphics[width=\textwidth]{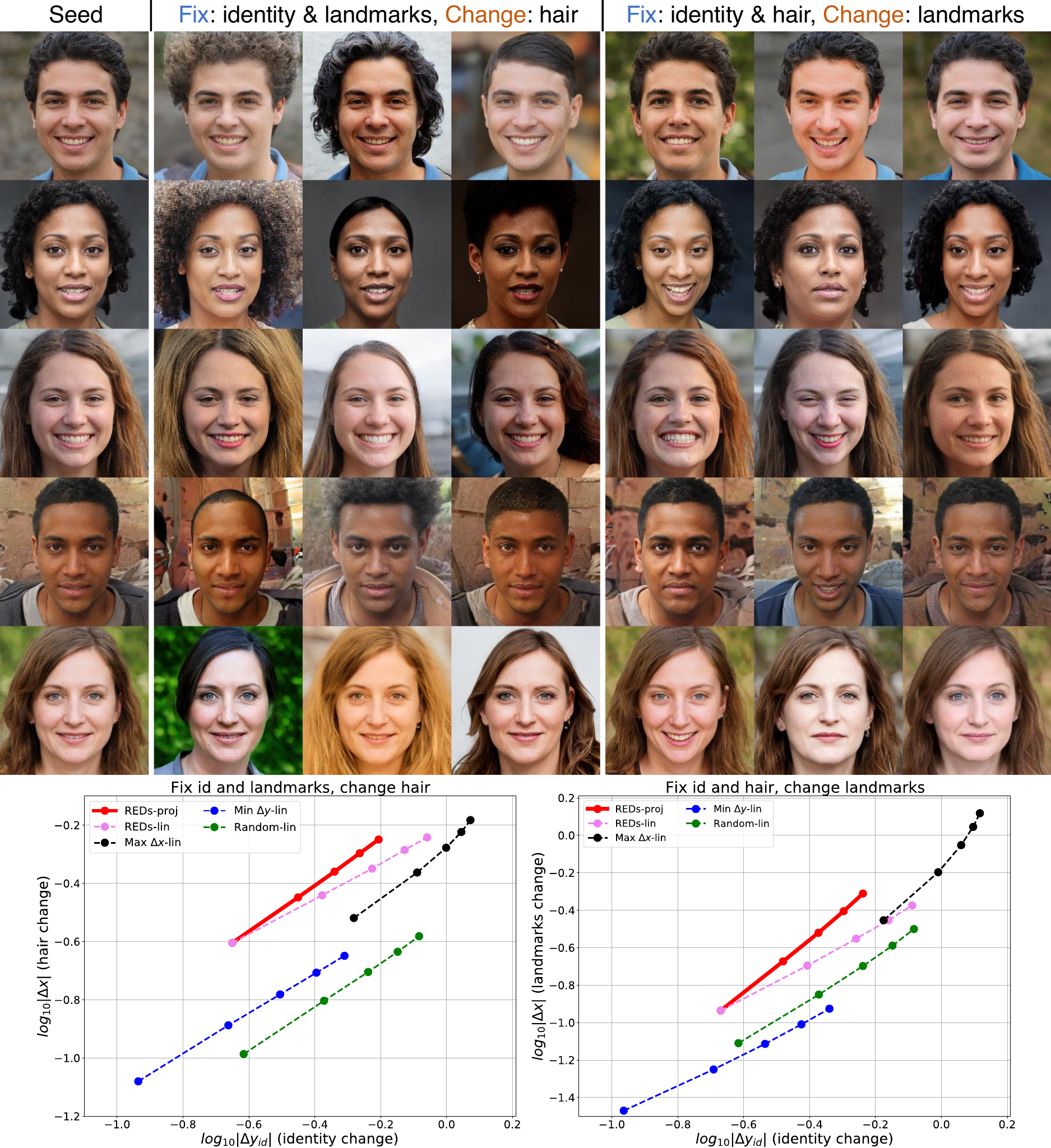}
    \caption{\textbf{Results for traversals controlled by identity, facial landmarks, and hairstyle.} (Top) Results using our method (REDs + projection) for 5 seed faces and two experiments: changing hair while fixing identity and landmarks (columns 2-4) and changing landmarks while fixing identity and hair (columns 5-7). We selected three samples per seed from different trajectories to illustrate the perceptual diversity of faces generated by our method while adhering to the fixed attribute constraints. (Bottom) Quantitative comparison of traversal methods. We generated $5$ traversals with $L=5$ steps for each method for 50 random seeds. We plot changes to hair (left) and landmarks (right) versus changes to identity in log-log scale, where each dot in the plot is the average value for each step over all examples. \textbf{\emph{Leftward and higher values are better}}. Our method using linear traversal (REDs-lin) outperforms the baselines also using linear traversal. Our method with projection traversal, REDs-proj, outperforms REDs-lin by reducing identity changes with no impact to hair or landmarks.}
    \label{fig:id-hair-landmark-samples}
\end{figure*}

\begin{figure}[t!]
    \centering
    \includegraphics[width=\linewidth]{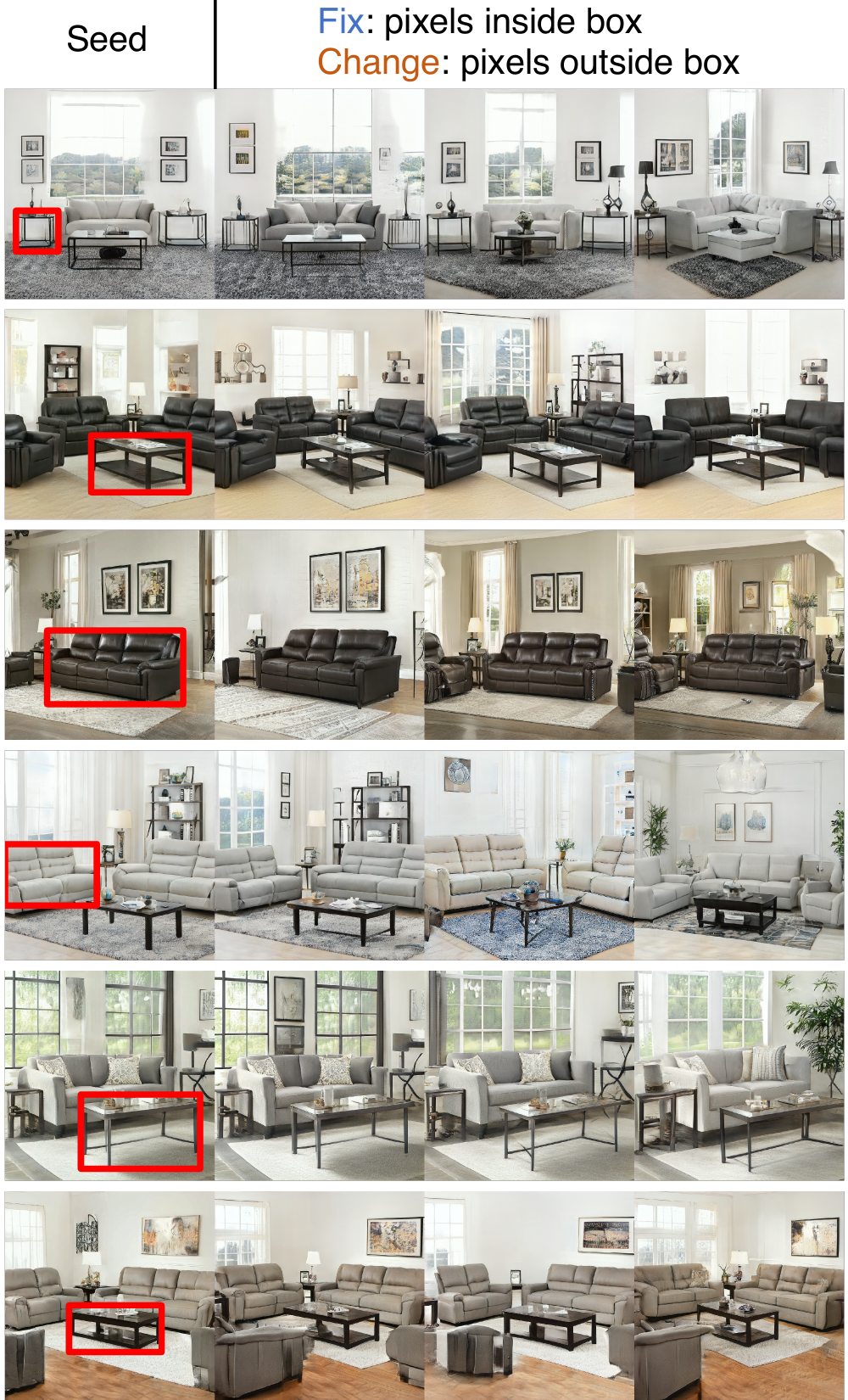}
    \caption{\textbf{Object-preserving living room traversals.} We used REDs with \emph{Projection} traversal, with $f(\cdot)$ and $c(\cdot)$ encoding raw pixel values inside and outside a bounding box on a piece of furniture (red box on seed image at left). The object within the box often stays fixed, but can undergo stylistic changes and movements (examples in rows 1, 4, 6) due to feature correlations in latent space. There are diverse changes to the rooms outside of the boxes, including new furniture (rows 1, 3, 4, 6), wall and window properties/decorations (all rows), and house plants (rows 4, 5).} 
    \label{fig:livingroomtraversal}
\end{figure}

\begin{figure*}[t!]
    \centering
    \includegraphics[width=\linewidth]{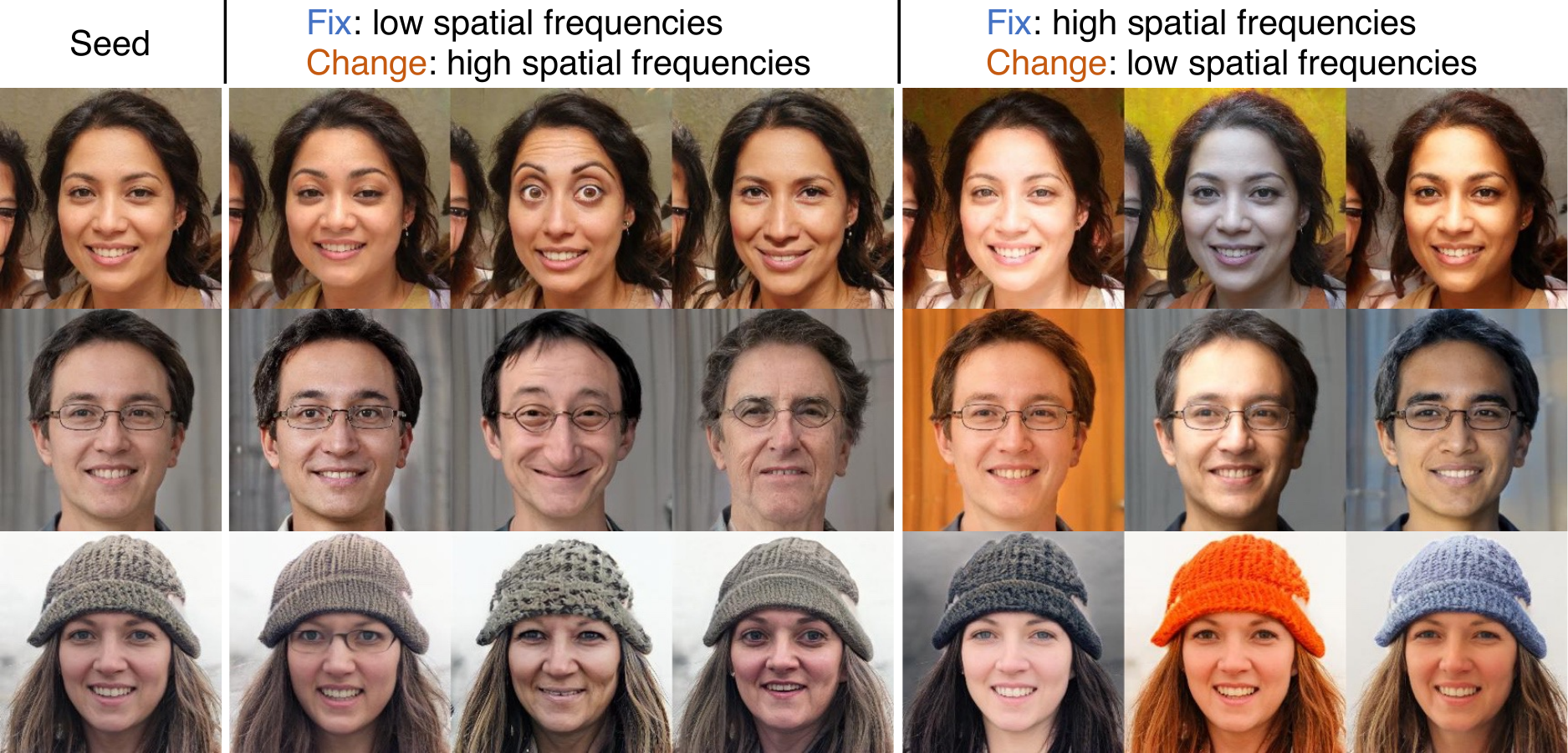}
    \caption{\textbf{Samples from traversals controlled by spatial frequency bands.} (Columns 2-4) The embedding function $f(\cdot)$ returns the raw pixels of the low-pass filtered image and $c(\cdot)$ the high-pass one. High-pass modifications change physiognomies, age and expressions, as well as hair and accessory texture, while the silhouette, lighting and color scheme are preserved. (Columns 5-7) $f(\cdot)$ and $c(\cdot)$ are inverted; low-pass modifications change colors, lighting and shading while mostly preserving identity, hair and textures.}
    \label{fig:freq-results}
\end{figure*}

\begin{figure*}[t!]
    \centering
    \includegraphics[width=\linewidth]{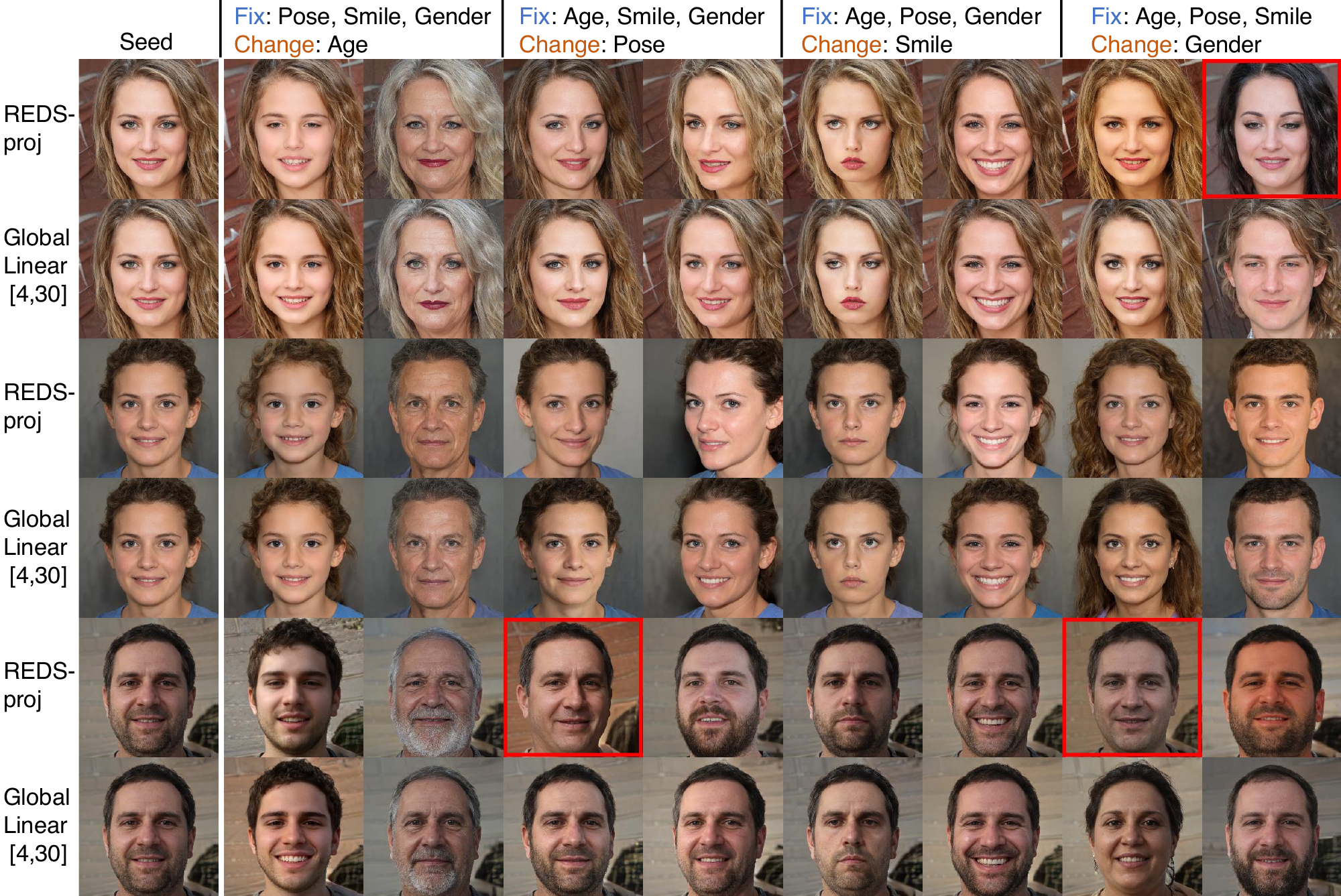}
    \caption{\textbf{Samples from traversals controlled by scalar semantic attributes.} On scalar attributes one may compare our method to a baseline of using a global linear model (SVM or ridge regressor) in latent space~\cite{balakrishnan2020towards, shen2019interpreting} (the global method is not defined and cannot handle multi-dimensional attributes). We change one attribute (age, pose, smile, gender) at a time while fixing the other three. Both methods are comparable for many cases. REDs sometimes fails (red-boxed images), particularly for gender (see Sec.~\ref{sec:limitations} for further discussion). }
    \label{fig:attribute-samples}
\end{figure*}

\subsection{Identity, hairstyle and landmark face traversals}
We first demonstrate our method on controlling three multidimensional facial features: identity, hairstyle, and 3D facial landmark positions.

We use ArcFace~\cite{deng2019arcface}, a popular open-source face identification model that encodes identity with a $512$-dimensional vector. To encode hairstyle, we run a public face segmentation model~\footnote{https://github.com/zllrunning/face-parsing.PyTorch} on each image, set pixels outside of the hair region to $0$, and flatten all pixels into a $256\times 256 \times 3 = 196,608$-dimensional vector. We encode 3D landmarks using the MediaPipe mesh model~\cite{mediapipe}, which predicts 468 landmarks around the face. This results in a $468\times 3 = 1404$-dimensional vector.

We performed two experiments: changing hair while keeping identity and landmarks fixed, and changing landmarks while keeping identity and hair fixed.  Fig.~\ref{fig:id-hair-landmark-samples} presents sample results for five test seed points using REDs and \emph{Projection} traversal. In both experiments, we set $\beta$s for fixed attributes to 0.99, and $\beta$ of the changing attribute to 0.999. We set both the Jacobian finite difference step and path step $s$ to $1$. We set path length $L=5$. Along with changing the input images along the intended features, our method is able to produce a \emph{wide variety} of different samples from different paths.

We quantitatively evaluated REDs against three baseline direction-finding approaches: choosing directions at random (\textbf{Random}), choosing the most significant eigenvectors of $A_c$, thereby maximizing changes to $\bx$ (\textbf{$\text{Max-} \Delta \bx$}), and choosing the least significant eigenvectors of $A_f$, thereby minimizing changes to $\by$ (\textbf{$\text{Min-} \Delta \by$}).

The plots in Fig.~\ref{fig:id-hair-landmark-samples} present our results. When using \emph{Linear} traversal, REDs outperforms the three baseline direction-finding approaches. $\text{Max-} \Delta \bx$ finds directions that significantly change hairstyle/landmarks and identity, $\text{Min-} \Delta \by$ preserves identity but also minimally changes hairstyle/landmarks, and \text{Random} performs worst of all. The figure also shows that when using REDs, \emph{Projection} outperforms \emph{Linear}. See Fig.~\ref{fig:strip-example} for a visual sample of this comparison and Supplementary for complete traversals.

\subsection{Frequency band face traversals}
Our method can handle arbitrary low-level image representations. We demonstrate this by controlling specific spatial frequency bands in Fig.~\ref{fig:freq-results}. We let $f(\cdot)$ and $c(\cdot)$ encode the raw pixels of low-pass and high-pass filtered versions of the input image (and vice versa). High-pass modifications change physiognomies, expressions and accessory textures. Low-pass modifications mainly change colors, lighting and shading.

\subsection{Object-preserving living room traversals}
We next apply our method to living room scenes. We aim to keep selected furniture fixed while changing other parts of the scene. We generated furniture bounding boxes with an object detector. We let $f(\cdot)$ encode the raw pixels within the bounding box, and let $c(\cdot)$ encode all remaining pixels in the scene. We set $\beta = 0.99$ for both features, a Jacobian finite difference step of $0.75$, path step $s=0.25$, and a path length $L=10$.

Fig.~\ref{fig:livingroomtraversal} shows several sample sequences. See caption for a detailed description. In Supplementary, we show sample strips of full traversals. We observe two notable degradations in these strips the farther we move away from the seed image. First, the `fixed' object often moves slightly at each step. Second, artifacts become more prominent because we rapidly advance to low-probability regions of the latent space.

\subsection{Scalar attribute face traversals}
\label{sec:scalar-experiments}
For scalar attributes, we can compare our method against a baseline that uses global linear directions~\cite{balakrishnan2020towards,shen2019interpreting}. These methods train a linear model per attribute (regressor for a continuous attribute or an SVM for a binary attribute) to predict the attribute value from the latent code. We change an attribute by moving along the hyperplane's normal direction. To fix other attributes, we orthogonalize the changing attribute's direction with respect to the other attribute directions.

Fig.~\ref{fig:attribute-samples} presents our results for four attributes: age, pose, smile, and gender. Overall, REDS-proj achieves similar qualitative performance to the baseline for most samples, but also has more failures cases when changing an attribute like gender, which often does not have a large local gradient in latent space. We discuss this more in~\ref{sec:limitations}. 

\section{Discussion}
Our experiments demonstrate the effectiveness of REDs at finding locally optimal orientations. By contrast, selecting random traversal directions or local directions that prioritize only one of Eq.~\eqref{eqn:obj} or~\eqref{eqn:constraint} do not work well due to the high dimensionality of the latent space (see plots in Fig.~\ref{fig:id-hair-landmark-samples}).

The superiority of \emph{Projection} over \emph{Linear} traversal (Fig.~\ref{fig:id-hair-landmark-samples}) also demonstrates the need for localized approximations of latent space geometry for complex image features. This is in contrast to past traversal studies~\cite{balakrishnan2020towards, ganspace, Jahanian2020On, Plumerault2020Controlling, shen2019interpreting} that found global linear directions to suffice for simple scalar attributes.

A consideration in all image synthesis works is the balance between perceptual quality based on human judgment, and quantitative optimization and analysis. In the application of faces, the user may have his/her own internal tradeoff curve between identity preservation and image diversity. Our method offers a principled way to explore different points on this curve by tuning the $\beta$ parameters (see Supplementary). Image perception also factors into the embedding functions used to measure image changes. 

GAN latent spaces are not all alike, and each requires different considerations. Faces are easier to model than living rooms, because the latter are a composition of many discrete objects interacting with one another. As a result, we found the face latent space and traversals to be smoother. Our living room traversals often exhibit large perceptual ``jumps'' due to discontinuities in latent space (see Supplementary). The complexity of a distribution also affects the degree of correlation between attributes. As Fig.~\ref{fig:livingroomtraversal} shows, it is not always possible to exactly fix a particular region of a living room while obtaining enough diversity elsewhere due to entangled features. Different regions of the latent space are also not alike. We found that high-likelihood regions produce the most realistic images and diverse traversals. Thus, the biases of the generative model have a direct effect on how well our method performs for a given image (see Sec.~\ref{sec:limitations} for further discussion). 

\subsection{Limitations}
\label{sec:limitations}
Our method takes a local view of the latent space to identify good traversal directions. However, as our results in Sec.~\ref{sec:scalar-experiments} suggest, there are benefits to taking a global view. Global linear models are likely better for attributes that are discrete, such as `wearing eyeglasses,' or approximately discrete for a large majority of samples like gender. For such attributes, local gradients in latent space can be near zero and swamped by noise. Another limitation of a local view is that gradients are undefined near sharp discontinuities in the latent space. We did not find this to be a decisive issue for faces, but did notice perceptual `jumps' in the living room scenes during traversals (see Supplementary for traversal strips). However, we note that our framework could be extended to use both global and local directions per traversal step, which we leave for future work.

Though our method can theoretically work with any deep generative model latent space, we used StyleGAN2 in all our experiments. Further experiments using other GAN or VAE architectures can give a more complete picture of our method's benefits and limitations.

\subsection{Ethics}
\textbf{Fairness}: As in past work~\cite{balakrishnan2020towards} we observed bias in StyleGAN's face distribution: Caucasian faces are most likely to be generated. This bias also affects trajectory quality, with light-skinned seed faces producing more diverse trajectories than dark-skinned ones. Biases in fixed and changing functions that use learning models also affect results. One example are face recognition models, like the one we used in our experiments to fix identity, which are known to have gender and ethnicity biases. To reduce bias one will want to train GANs and any learned models on rich and diverse datasets.

\textbf{Fake portrayals}: GANs could be used to generate fake images of individuals under different conditions. This could include the case where the image of the face of a real person is projected onto the GAN latent space and then manipulated.

\section{Conclusion}
We presented a simple, principled and versatile method designed to explore a generative model's latent space to produce sets of synthetic samples where one group of multidimensional features is held constant while another is varied as much as possible. We demonstrated traversal results on several features that previous works are not capable of handling: landmark locations, pixels within regions, frequency information, and facial identity as measured by a deep neural network. Our experiments show the need for modeling local geometry of latent spaces for high-dimensional features. Understanding the complex nature and geometry of the latent space of image generators is a fascinating question which we have only started to explore.

{\small
\bibliographystyle{ieee_fullname}
\bibliography{egbib}
}

\end{document}